\documentclass{article}
\usepackage{ijcai27}
\usepackage{times}
\usepackage{xcolor}
\usepackage{fontawesome5}
\usepackage{url}
\usepackage[hidelinks]{hyperref}
\usepackage[utf8]{inputenc}
\usepackage[small]{caption}
\usepackage{graphicx}
\usepackage{amsmath}
\usepackage{booktabs}
\usepackage{array}
\title{ComputeWise: Measuring and Fixing GPU Waste by AI Research Agents}
\author{
Egor Pakhomov$^1$\thanks{\raggedright Lead author. Correspondence to: pahomov.egor@gmail.com.}, Nina Pakhomova$^2$, Anurag Koul$^1$, Shafiq Joty$^1$ and Erik Nijkamp$^1$
\affiliations
$^1$Salesforce AI Research\qquad $^2$Datadog\\[6pt]
{\normalsize\definecolor{relink}{RGB}{0,84,166}%
\href{https://github.com/SalesforceAIResearch/computewise}{\raisebox{-0.12em}{\large\faGithub}\,\textcolor{relink}{\textsf{SalesforceAIResearch/computewise}}}\hspace{2.2em}\href{https://huggingface.co/datasets/Salesforce/computewise}{\raisebox{-0.22em}{\includegraphics[height=1.15em]{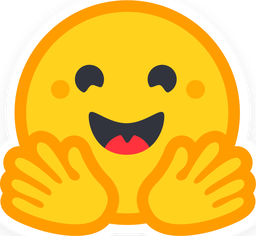}}\,\textcolor{relink}{\textsf{Salesforce/computewise}}}}
}

\begin{document}
\maketitle
\begin{abstract}
AI agents are given GPUs to do AI research on their own. Do they use those GPUs well? A busy GPU does not tell us, because it can be busy with work the answer did not need. This matters, because the tokens the agent spends cost almost nothing next to the GPUs its experiments run on, so using those GPUs better has a huge impact. So we built ComputeWise, an open benchmark of research tasks on which this waste can be measured and reduced. For it we developed our own simulated GPU, which replays recorded runs to the agent's code. With no real GPU needed, we ran hundreds of experiments, far more than would be reasonable on real GPUs. With no help, the agents put only 52\% of their GPU time into work the answer needed. Another 22\% went to work it did not need, and 26\% sat idle. Asking the agent to finish the work faster, telling it to keep every GPU busy, or giving it a job scheduler does not fix this. So we built an open-source harness, part prompting and part deterministic code. Across five models, it removed over 80\% of the extra time to the answer.
\end{abstract}
\section{Introduction}
\label{sec:intro}

AI agents now run machine-learning experiments on their own. Given a question and a few GPUs, an agent writes the scripts, starts the training runs, reads the results and decides what to run next. Benchmarks for these agents score the result they reach \cite{mle-bench,re-bench,mlagentbench}. They do not ask how well the agent used the GPUs it was given. That question matters, because the tokens an agent spends cost almost nothing next to the GPUs its experiments run on.

\begin{figure}[t]
\centering
\includegraphics[width=\columnwidth]{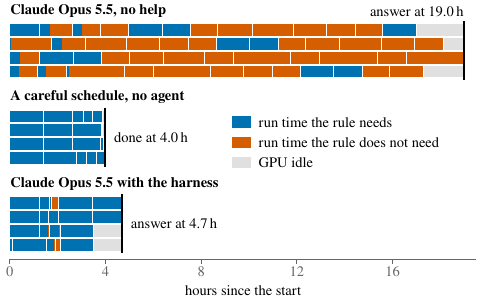}
\caption{One task, three ways. The question: which of nine optimizers to adopt. Each strip is four GPUs. Each block is one training run, coloured by whether the task's rule needs it. Grey is a GPU left idle before the answer (black line). With no help, Claude Opus~5.5 ran all 50 runs the files list. The rule needs ten of them in full and the first 3 to 27 minutes of eight more. Giving every candidate all five runs, but stopping each run where the rule decides it, would take 6.5 hours by our calculation. Each agent run sits at the median of its group: three runs with no help, six with the harness. All three strips ran on one version of the simulated GPU.}
\label{fig:onetask}
\end{figure}

Earlier studies have seen agents leave GPUs idle \cite{mle-bench,agent4sci}. The usual measure is utilisation, the share of time the GPUs are busy \cite{enpire,athanor}. But a busy GPU is not always a useful one, and Figure~\ref{fig:onetask} shows why. We asked Claude Opus~5.5 to choose among nine candidate optimizers for training a language model. An optimizer should be adopted only if it trains fast enough and reaches a lower loss than our baseline. Eight of the nine show in their first half hour of training that they cannot pass. A careful engineer would stop those eight right there and finish only the runs that still matter, which takes 4.0 hours. Opus instead trained every optimizer to the end, fifty runs in all. Its answer was right, but it took 19.0 hours, with the GPUs busy 93\% of the time. The run's tokens cost about \$2, and the GPU time it held costs about \$520 at cloud prices.

So there are two kinds of waste. One is idle GPUs. The other is unneeded work: runs that the task's own rule did not need for the answer. Utilisation sees the first kind and counts the second as success.

So we built ComputeWise, an open benchmark of research tasks on which this waste can be measured and reduced. Its measure is wall-clock time to a correct answer. That is the right measure because the GPUs are reserved for the agent until it answers: idle, on unneeded work or on useful work, they cost the same. We report it as a ratio to a careful schedule for the same task and GPUs, so 2.0 means the agent took twice as long. The run in Figure~\ref{fig:onetask} reads 4.8.

Running these experiments on real GPUs would be very expensive: a single run can hold four GPUs for most of a day. So we developed our own simulated GPU. It replays recorded runs from public sources \cite{mls-bench,modded-nanogpt} to the agent's code: an eight-hour job still takes eight hours and returns the numbers that were recorded. The numbers matter as much as the timing, because the agent decides what to run next from them. Earlier GPU emulators reproduce the timing, but not the numbers \cite{revati,phantora,maya-gpu-emulation}. Because the numbers are recorded, the right answer to each task is known in advance, so agents differ only in how long they take to reach it. And because no real GPU is needed, we ran hundreds of experiments, far more than would be reasonable on real GPUs.

With no help, eight models from top AI labs put only 52\% of their GPU time into work the answer needed. Another 22\% went to work it did not need, and 26\% sat idle. One run in four took at least twice as long as a careful schedule, and one in fourteen took more than four times as long. Asking the agent to finish the work faster, telling it to keep every GPU busy, or giving it a job scheduler does not fix this. With the last two, the share of GPU time spent on work the answer did not need even went up, from 21\% to 26\% and 25\%. Other researchers found the same: reminding agents to use their hardware lowered their scores \cite{how-much-harness-mle}, and a tool for running jobs in parallel made results worse \cite{researchgym}.

So we built an open-source harness, part prompting and part deterministic code. The prompting part has the agent write a plan before anything runs: which jobs the question needs, and which early results would make other jobs unnecessary. The code part then runs the plan: it starts the jobs in a good order, stops the ones no longer needed, and wakes the agent only when a result needs judging. Across five models, it removed over 80\% of the extra time to the answer. For four of the five, a run costs under \$15 in tokens and frees up to 57 GPU-hours.

Our contributions are:

\begin{itemize}
\item \textbf{ComputeWise}, an open benchmark of research tasks that measures an agent's wall-clock time to a correct answer, not how busy its GPUs are.
\item \textbf{A simulated GPU} that replays recorded runs to an agent's own code with their recorded timing and results, so agents can be tested with no real GPU.
\item \textbf{A study of eight models from top AI labs.} With no help they spend 22\% of their GPU time on work the answer did not need and leave another 26\% idle, and simple fixes do not remove this.
\item \textbf{An open-source harness} that removed over 80\% of the extra time to the answer across five models.
\end{itemize}

\section{Related Work}

Idle accelerators already appear in evaluations of research agents. MLE-bench found no attempt to use an added GPU in the logs inspected; RE-Bench reports little background execution despite reminders \cite{mle-bench,re-bench}. Agent4Sci likewise describes idle GPUs and experiments run serially when they could overlap \cite{agent4sci}. The AI Scientist reports low utilisation for its small, non-compute-intensive experiment templates \cite{ai-scientist}. Benchmark outcomes include medals in MLE-bench, scores across time budgets in RE-Bench, and task success alongside wall-clock time and token consumption in MLAgentBench \cite{mlagentbench}.

ENPIRE reports time to success as well as the fraction of research wall-clock time for which GPUs are active; no evaluated coding agent reaches full utilisation \cite{enpire}. In Athanor's 104-hour post-training campaign, jobs held GPUs for 93\% of available GPU-hours, while device-reported utilisation was about 75\%; less than 1\% of GPU-hours passed with an idle GPU waiting for its agent \cite{athanor}. Allocation, activity and waiting for an agent are separate quantities even within the same campaign.

Attempts to improve resource use have mixed outcomes. On single-GPU ML engineering, reminders triggered by low CPU or GPU utilisation lowered MLE-bench scores; removing background-job tools made no significant difference \cite{how-much-harness-mle}. ResearchGym's parallel-job tool introduced coordination failures and worsened results \cite{researchgym}. Conversely, Agent4Sci improved research scores with a bundle of written skills covering ideation, experiment management and periodic review; the experiment-management contribution was not isolated. AOS-Bench compares dispatch policies for CPU workloads, checking completed outputs and recording core utilisation; in its policy ablation, median CPU use stayed below 30\% under every policy while tasks waited \cite{aos-bench}. Other systems put dispatch in code: AlphaLab assigns freed GPUs to queued experiments, while AIRA$_2$ dispatches new work as dedicated GPU workers become available \cite{alphalab,aira-2}.

Systems research has long distinguished productive computation from occupied hardware. ML Productivity Goodput separates scheduling, runtime and program efficiency \cite{ml-fleet-efficiency-goodput}. In a production GPU cluster, Jeon et al.\ \shortcite{philly-gpu-cluster-analysis} find that passed jobs spent 62\% of their GPU time, on average, after coming within 0.1\% of their lowest training loss. For research agents on tabular problems, Au et al.\ \shortcite{recovering-wasted-compute-autoresearch} find that 46\% of baseline search nodes in AIDE, a tree-search agent, re-encountered a bug already seen in the same run.

TimeArena evaluates completion time in simulated cooking, household and laboratory activities against a manually calculated greedy reference, and distinguishes necessary from unnecessary waits \cite{timearena}. For machine learning, NAS-Bench-101 lets optimisers query recorded accuracies and training times; HPOBench provides tabular and surrogate objectives for hyperparameter optimisation \cite{nas-bench-101,hpobench}. Syne Tune adds a simulated clock and ordered result events for asynchronous workers \cite{syne-tune}. Its authors note that real start-up and shutdown delays can make frequently launching fresh trials less attractive.

GPU emulators preserve the running program instead: Revati advances virtual time while serving code executes; Phantora profiles kernels on a real GPU and advances per-process clocks; Maya captures a device-operation trace and simulates its timing afterwards \cite{revati,phantora,maya-gpu-emulation}. These emulators model runtime without computing the program's numerical results. They therefore support performance estimation, but do not supply meaningful losses or scores to code whose next action depends on them.

\section{The Benchmark: From Job Lists to Research Tasks}
\label{sec:benchmark}

We built the benchmark twice: first as lists of jobs to run, then as research questions to answer.

\subsection{The Simple Suite}

The simple suite has six shapes of job list: independent jobs, jobs of uneven length, one long job among short ones, a chain of checkpoint resumes, scripts that must first be written, and screening runs whose best are continued. The jobs replay recorded training runs of a small GPT at their recorded pace (Section~4). A brief names them and the four GPUs, asks for every result, and says nothing about time. We ran Claude Haiku~5.5 and Claude Sonnet~5.5 in Claude Code and GPT-6 Astra in Codex, with no help and with two fixes: a \emph{hint sentence} added to the brief, asking that no GPU sit idle while work remains, and a job \emph{scheduler} inside the machine, which the brief says to use. A run that answered is read as its time to the last job over the \emph{best computed schedule} of its jobs, the shortest of several computed knowing every job's length.

Given fourteen jobs of 6 to 27 minutes, Sonnet dealt them out to the four GPUs in advance, in list order, in all eight no-help runs. Three GPUs ran out of work while jobs waited behind the fourth, and every run took 1.45 times the best schedule. With the hint its median fell to 1.02, and with the scheduler to 1.08 (two runs each). On the short lists, seven pairs of a model and a list had a no-help median of 1.25 or more, from 1.39 to 2.95; the better of the two fixes brought each to between 1.01 and 1.16, some on a single run.

The suite could not be the benchmark: on its short lists a sentence or a scheduler removes the waste, and a list cannot show unneeded work, because it asks for every job.

\begin{figure}[!t]
\centering
\includegraphics[width=\columnwidth]{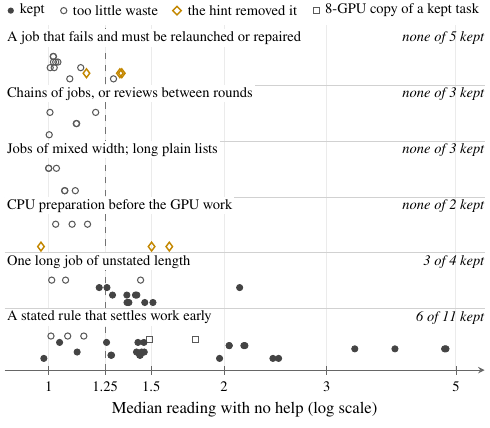}
\caption{The 28 designs screened, grouped by how each was expected to cost an agent time. A mark is one model's median no-help reading on one design, over one to seven runs that answered. The nine tasks kept (filled) have at least two models at or right of 1.25 (dashed) because the conditions required it; on the two KV results tables that depends on what the comparison schedule knows (Table~\ref{tab:nine}). A dropped design's mark is its stored ratio to its own comparison, on some designs to the end of the work and not to an answer; the four designs without supported answers have none. The twelve dropped for too little waste were judged on 85 runs, 6 to 12 a design: that is weaker than showing they have none.}
\label{fig:designs}
\end{figure}

\subsection{Twenty-Eight Designs}

Of more than forty public sources of recorded runs, four entered our registry, with 698 runs. The nine tasks replay two: MLS-Bench's recorded baseline evaluations \cite{mls-bench} and optimizer training runs from the modded-nanogpt repository \cite{modded-nanogpt}. We built 78 designs, the suite's lists among them; 28 were realistic research requests, each built on one expectation of how agents lose time (Figure~\ref{fig:designs}).

Six conditions, fixed before any screening run was read, decided which stayed. A design needed a brief with no procedure or word about time, and a \emph{reference schedule}, one we wrote to follow the brief's rule and ran with no agent, that finishes within 10\% of the best computed schedule. Agents had to reach supported answers (four designs failed), and with no help their median reading had to be 1.25 or more for at least two of three models (twelve failed). With the hint, then with the scheduler, at least one Claude model had to stay at 1.15 or more and keep half its excess over 1 (two failed, both with the hint). The nineteenth dropped design was an eight-GPU copy of a kept task.

Most expectations were wrong. Given a chain of five training pieces that cannot overlap, listed after 36 independent runs, all four no-help runs started the chain in under a minute.

Two kinds of design held. In one, every job is needed, and one job far longer than the rest has a length no file gives. In the other, the brief states a rule under which an early result settles a candidate, and its remaining runs become unneeded work. The nine were selected for waste that persists; how often they show it does not estimate how often research does.

\subsection{The Nine Tasks}

A task (Table~\ref{tab:nine}) is a one-paragraph brief, scripts and data, and a machine with four simulated H100 GPUs (eight on one task). Every job uses one GPU and replays one recorded run at its recorded pace, so the right answer is fixed in advance; the runs were recorded on several GPU types, mostly H100s.

\begin{table*}[t]
\centering
\includegraphics[width=\textwidth]{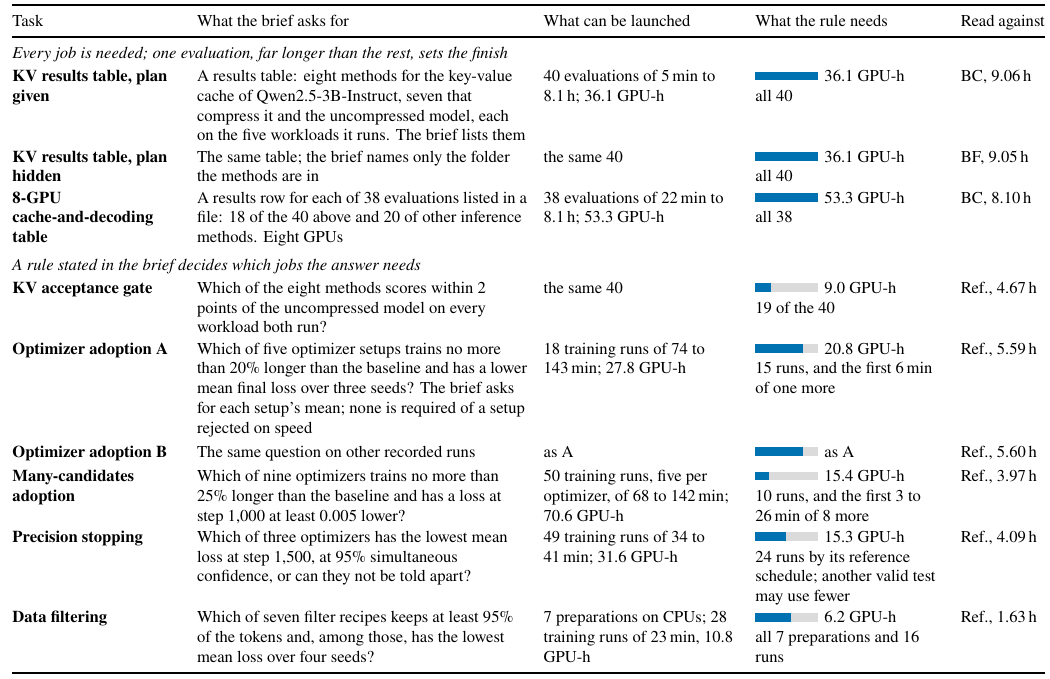}
\caption{The nine tasks. ``What can be launched'' counts every job the files offer, run once and to its end. ``What the rule needs'' is the work the comparison schedule runs; the bar is its share of the GPU-hours that can be launched. Launching everything, with no GPU ever idle, takes at least 1.24 times the comparison schedule on optimizer adoption A and B, 1.66 on data filtering, 1.93 on precision stopping and the acceptance gate, and 4.44 on many-candidates adoption. Every job uses one GPU; lengths are at the recorded pace. ``Read against'' is the comparison schedule a reading divides by, with its time: Ref., the reference schedule; BC, the best computed schedule; BF, the best feasible schedule (Section~4). The comparison knows more than the brief gives. On the two KV results tables it has the evaluations' lengths, which no file states. On the simulated GPU a trial of two examples of each evaluation recovers their order at a cost of 0.3\%, exactly so only because every example is paced alike; against a schedule ordered by workload size, with no trial, three of the six Claude medians on those two tables would fall just under 1.25. On many-candidates adoption the reference schedule knows which candidate passes; without that, readings there would be about 9\% lower.}
\label{tab:nine}
\end{table*}

Forty recorded evaluations underlie four of the tasks and show both kinds. They score eight methods for a language model's key-value cache, the uncompressed model among them, on five workloads each. One takes 8.1 hours, where one GPU's share of all forty is 9.0. A results table needs all forty. The best computed schedule starts the long one first and ends after 9.06 hours; the brief names it 36th of 40. When the brief asks instead which method stays within 2 points of the uncompressed model on every workload, the reference schedule runs 19 of the forty and never the long one, whose method fails on a 22-minute evaluation.

The other five replay training runs. Optimizer adoption~A and~B ask one question on separate recorded runs: one of five candidate setups fails the rule on speed, six minutes into its first run. In many-candidates adoption, a log line rules out eight of nine candidates within half an hour of a first run, and the rule needs 15.4 of the 70.6 GPU-hours on offer. Precision stopping leaves the number of runs to the agent: its reference schedule tests at three points fixed in advance and stops at the second, after 24 of 49 runs. In data filtering, seven recipes, six built from FineWeb's filters \cite{fineweb} and one applying none, run for real on the CPUs; three keep too few tokens, and their twelve training runs are unneeded.

The nine are not nine independent samples: three share the forty evaluations, a fourth reuses 18, and five draw on one repository. Every rule and threshold is ours, and the filtering task's losses are recorded runs we assigned to the recipes, not effects of the filters. Three briefs were reworded once after their first runs.

\section{Replaying the Work, Timing the Answer}
\label{sec:measure}

An agent's script starts as a real process: imports, data loading and tokenisation run on the CPU. Patched PyTorch matches GPU work by script hash, arguments and environment settings; records may also permit path or checked configuration matches. Matching work waits at the recorded pace and returns stored values; unmatched edits run on the CPU or are refused. The simulated GPUs report H100s to monitoring tools. Whole evaluations return recorded scores; partial evaluations include constructed scores, on which no task's answer depends. Deliberation, commands and waiting take real elapsed time. No agent ran on physical GPUs.

The in-machine scheduler implements SLURM submission, dependencies, arrays and cancellation. It starts jobs in submission order, fitting later jobs into available resources. The agent chooses the work, which uses the same replay.

The \emph{reading}, or time to the answer, divides time from agent start by the \emph{comparison schedule}'s time on the same task and GPUs. A reading of 1 matches that schedule; Table~\ref{tab:measurement-clocks} gives the endpoints. Time before the first launch counts.

\begin{table}[!htb]
\centering
\includegraphics[width=\columnwidth]{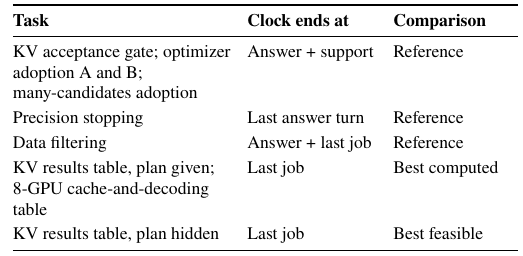}
\caption{All clocks start with the agent. A ``+'' means the later event; support ends with required work. Results-table answer waits are reported separately.}
\label{tab:measurement-clocks}
\end{table}

We wrote the \emph{reference schedule} to follow the task's rule; it runs without an agent and ends with its last job, checks included. It knows lengths absent from the brief and, on many-candidates adoption, which candidate passes. Section~3 quantifies these advantages. The \emph{best computed schedule} is the shortest of several computed knowing the agent's own jobs and their lengths, respecting when each can start. It measures their arrangement, absorbing added work into the comparison. The \emph{best feasible schedule} fixes all recorded evaluations, one per GPU, independently of the agent's launches. Neither label claims a proven optimum; an agent can beat a comparison.

A supported answer must match the key and have the required evidence. Wrong, unsupported or absent answers have no reading; other exclusions include simulator faults. One incomplete no-help answer retains a marked lower bound; harness lower bounds are excluded from medians.

Unneeded work is work the task's rule does not require. GPU time held is GPU count times the interval from start to the later of answer and last job. The headline split counts busy time beyond the reference's busy-time budget as unneeded; the rest is needed or idle. These shares do not divide elapsed delay into causes.

Both times use the same records, so uniform timing errors nearly cancel; agent delays do not scale. We checked 194 jobs on eight H200s, the available hardware, covering setups from seven tasks; data filtering and part of the 8-GPU table went unchecked. All 14 training time-rule decisions matched their records. Under the records' PyTorch, 38 of 39 KV evaluations scored within 1.6 points of their records and the gate's decision matched; a newer PyTorch changed it. Re-timing the gate with real job lengths moved four models' no-help medians 1--4\% (two or three runs each).

In two pairs of small evaluations sharing a GPU, each job retained 82--89\% of solo speed, against 50\% simulated. Re-timing three no-help runs per model on the KV results table with the plan hidden lowered medians from 1.29 (range 1.14--1.38) to 1.14 (1.06--1.29) for GPT-6 Astra in Codex and from 1.41 (1.37--1.61) to 1.23 (1.16--1.29) for Claude Opus~5.5. Both fall below 1.25, the waste threshold fixed before any screening run was read. These are modelled re-timings of fixed agent choices.

The simulator changed during the study. On five tasks version and condition are partly confounded: the harness used a later version than most no-help, hint and scheduler runs. Matched jobs retained their times, but launch matching, partial scores and visible signs of simulation changed. Readings affected by earlier CPU fallbacks are retained, with estimated changes bounded by 0.03.

\section{Where the Extra Hours Go}
\label{sec:waste}

Across the nine tasks, 52\% of held GPU time went to needed work, 22\% to work the answer did not need, and 26\% was idle (pooled over 148 runs; Table~\ref{tab:nohelp-cost}). Of the 125 runs with a reading, 32 took at least twice as long as the comparison schedule, and 9 more than four times as long. Only 5\% was idle beside a ready job, over 146 runs with detailed logs. Selection favoured that small share: tasks stayed only if waste survived the hint sentence and scheduler, both aimed at idle GPUs beside waiting work. These tasks therefore do not estimate how often research agents waste compute.

\begin{figure}[t]
\centering
\includegraphics[width=\columnwidth]{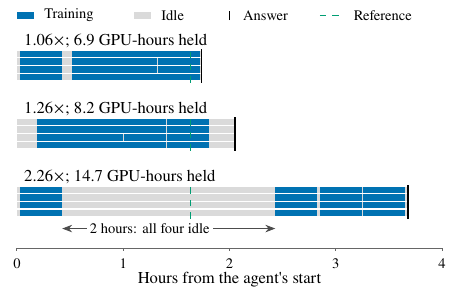}
\caption{All three no-help Opus runs on data filtering, ordered by reading. Each strip contains four GPU lanes; blue is training and grey is idle capacity held until the answer (black tick). Every run trains the same four usable recipes on four seeds. The slowest leaves all four GPUs empty for two hours between training waves. The dashed line marks the reference schedule's finish. The reading includes the wait for the answer.}
\label{fig:filter-three}
\end{figure}

Three runs by Claude Opus~5.5 in Claude Code on data filtering performed the same 16 training jobs and gave supported answers. The fastest held four GPUs for 1.74 hours; the slowest, 3.68 hours. That is 6.9 versus 14.7 GPU-hours of capacity for the same training work. Their \emph{readings}, time to the answer divided by the comparison schedule's time, were 1.06 and 2.26. Figure~\ref{fig:filter-three} includes the third run, at 1.26: its jobs finished, then it waited another 15 minutes to answer.

Here the rule that excludes work was followed: none of the 13 data-filtering runs trained a recipe that failed the token-retention check. What varied was when the allowed work started. Pooled over those runs, 25\% of GPU time was free while a training job could have started; for 24 percentage points the agent was sleeping or waiting. In the slow Opus run, preparation used a single CPU filtering task per recipe, and a two-hour gap opened between training waves. The agent's stated reason for that setting was ``so document order stays canonical''.

Table~\ref{tab:nohelp-cost} puts these runs beside the rest of the benchmark. Its readings describe supported answers, with one disclosed lower bound. Of 150 no-help runs, 25 have no reading: 13 failed to deliver; five received an earlier brief, three refused after concluding the GPUs were simulated, three were shaped by a simulator limit, and one obtained its answer from the simulator's interface. The GPU-time account includes runs without readings. It measures held capacity, not excess wall-clock hours. The prose classes jobs by the decision rule; across all tasks, its shares agree with the budget account within two percentage points. Repeated launches contribute to the small excesses on results tables.

\begin{table*}[t]
\centering
\includegraphics[width=\textwidth]{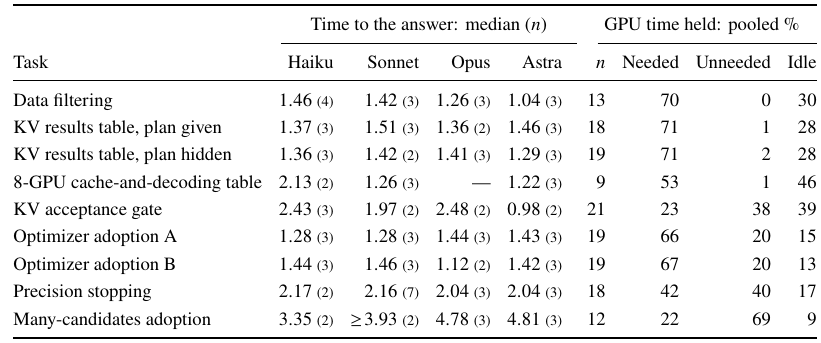}
\caption{No help: model-specific readings and the GPU-time account. Left: median reading (number of readings) for 35 model-and-task cells: Claude Haiku~5.5, Claude Sonnet~5.5 and Claude Opus~5.5 in Claude Code, and GPT-6 Astra in Codex. These four models have the widest no-help task coverage. A dash means no runs. Sonnet's many-candidates cell includes one lower bound; $\geq$ marks the resulting bound on its median. Results-table readings end at the last job; the other clocks follow Section~4. Right: percentages of capacity held, pooled over all models with a split ($n$ runs), including runs without readings. Busy time counts as needed up to the reference schedule's busy GPU time; the rest counts as unneeded. Idle is the remainder. The window ends at the later of the last job and the answer. Rounding may prevent totals of 100. Adoption A and B carry the brief-wording caveat below.}
\label{tab:nohelp-cost}
\end{table*}

\paragraph{A late job leaves a long tail.}
On the three results tables, every requested evaluation is needed. The loss is largely idle GPUs after their last job, while another GPU finishes. Of the 38 runs with readings, 32 started the long evaluation more than an hour in. Its median start over all 38 was 3.4 hours, taking each run's first launch that lasted at least an hour. On the plan-given KV results table, this tail occupies 25.5\% of capacity, pooled over 18 runs, against 0.2\% for the reference schedule. On the 8-GPU cache-and-decoding table the corresponding shares are 41.7\% over nine runs and 17.6\%. Some idle capacity is unavoidable even in a good schedule.

These comparisons benefit from knowing job lengths. As Section~3 reports, ordering by workload size instead would bring three of the six Claude medians on the two KV results tables below the 1.25 waste threshold, fixed before screening. A reading above that level is relative to the stated comparison schedule, not proof that the agent could infer its ordering for free.

\paragraph{A full queue can prolong the decision.}
Many-candidates adoption reverses the picture. Nine of twelve runs completed all 50 training runs. Across all twelve, 68\% of capacity went to candidates after their own results had ruled them out. This is \emph{unneeded work} under the task's decision rule, regardless of its other scientific uses. GPT-6 Astra's three readings were 4.55--4.81; its fastest run kept the GPUs 98\% busy and took 18.1 hours. The task files request five runs per setup. Even granting every setup five runs, but stopping each at its deciding log line, a computed schedule in file order takes 6.5 hours on four GPUs, without knowing job lengths. This calculation uses the recorded time to each stopping point, without delays for reasoning or issuing commands.

The reference schedule also knows which candidate passes. Without that knowledge, readings there would be about 9\% lower. This changes the scale modestly, while leaving the excess work in view.

Optimizer adoption A and B show a smaller version: the rejected setup's three runs all finished in 30 of 38 attempts. Their briefs ask for ``each setup's mean'', which can reasonably request that work. Rewording A to ask for ``the means your decision rests on'' gave median readings of 1.43 for Astra, 1.29 for Opus and 1.34 for Sonnet, two runs each. Excess time persisted; these small cells do not show that wording had no effect.

Precision stopping exposes a different ordering cost. Of capacity pooled over 18 runs, 22\% went to replications after the comparison was settled and 18\% to surplus replications of one method before the others supplied enough results. Extra work need not begin after a decisive result to be unneeded.

\paragraph{Success does not transfer uniformly.}
Astra's two readings on the KV acceptance gate were 0.97 and 1.00, yet its three precision-stopping readings were all 2.04. Within the Claude models, Opus has a lower median than Haiku on optimizer adoption B and a higher one on A. Small cells also hide wide spreads: Opus's two plan-given KV readings are 1.02 and 1.71. Table~\ref{tab:nohelp-cost} supports task-specific comparisons, not a model ranking.

Over 87 runs across the six decision tasks, time to the last job was a median 1.05 times the best computed schedule of the jobs launched; the median reading against task-specific reference schedules was 1.46. The ratios differ in work sets and clock endpoints, so their gap is not a decomposition of lost time. Packing was less close on data filtering and the KV acceptance gate: ratios of 1.17 and 1.25, respectively, over 13 runs each. The next sections test fixes against both delays and unneeded work.

\section{What the Simple Fixes Reach}
\label{sec:fixes}

The hint sentence asks that every GPU stay busy ``until all the runs are finished''; the in-machine scheduler starts submitted jobs on free GPUs in order. Claude Haiku~5.5 and Sonnet~5.5, in Claude Code, made 55 runs with the hint, 54 with the scheduler and 62 with no help. These nine tasks were kept only if waste survived both fixes: they show what the fixes change, not how often they suffice. We also tried a sentence asking the agent to get the work done as fast as it can, in 14 runs with Haiku and Sonnet on five of the tasks. Its eight readings run from 1.05 to 2.56, with a median of 1.43.

Each fix cut the waste it aims at. Pooled over GPU time (61 no-help runs have a split), a GPU sat free beside a launchable job 5.9\% of the time with no help, 0.5\% with the hint and 2.6\% with the scheduler, mostly where jobs' CPU requests capped how many ran, or training was submitted in batches. That was a fifth of the no-help idle time, small partly because selection dropped tasks the fixes repaired. GPUs idle with nothing to launch, mostly behind a late long job, held 22.4\%, 22.3\% and 21.5\%; unneeded work rose from 21.3\% to 26.3\% and 24.6\%.

Readings (time to a supported answer over the comparison schedule's) moved with what neither fix sets: which job starts first, which jobs run at all. On the results tables, where every job is needed, the reading turns on when one 8.1-hour evaluation's turn comes in the agent's list. Haiku's fix runs all started it 21st of 38 on the 8-GPU table, five at 2.1 hours against 5.3 to 5.8 with no help, and its typical reading fell from 2.13 to 1.26 (Table~\ref{tab:fixes}). Sonnet, already at 1.26, did not improve. On the KV tables, where it is 36th of 40 in the brief and 31st in the folder listing, its median start moved from 4.0 to 5.6 hours with the hint and 5.4 with the scheduler (twelve runs each); seven of eight fix cells read higher.

On two decision tasks more runs launched everything on offer, as the hint's last clause asks. On many-candidates adoption three of six no-help runs launched fewer than its 50 runs, one of them in each model's no-help median; under either fix all twelve launched 49 or 50, every typical reading rose, and the task holds most of the pooled rise in unneeded work. On optimizer adoption~A, whose brief can be read as asking for them, the rejected setup's three runs all went to the end in one of six no-help runs and eleven of twelve fix runs. Sonnet with the scheduler on precision stopping, the one decision-task cell to lose half its excess over 1 (2.16 to 1.45), rests on two runs that launched only 33 and 34 of the 49 runs; one no-help run launched 27.

On every task one model keeps at least half its excess under each fix, as selection required, so at most 18 of the 36 fix cells could lose half; three did, and ten hint and nine scheduler cells read higher than with no help.

\begin{table}[t]
\centering
\includegraphics{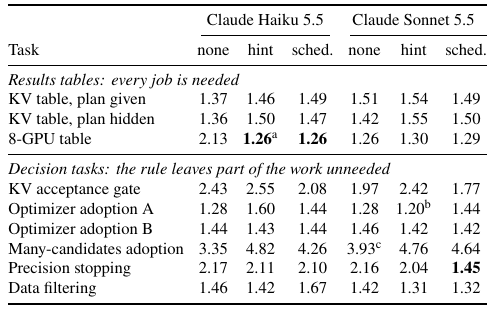}
\caption{Typical readings (the median of a cell's readings) of the 18 pairs of a task and a model under no help, the hint sentence and the scheduler. Cells rest on two to four readings, except Sonnet's no-help cell on precision stopping (seven) and the cell marked b (one). Bold: the fix left at most half of the pair's no-help excess over 1. Marked a: Haiku's hint runs here had a reworded brief; its no-help and scheduler runs share the original. Marked c: includes the one lower bound, 3.06, so this median is itself a lower bound. The optimizer adoption briefs ask for ``each setup's mean'', which can be read as asking for the rejected setup's runs; on adoption~A with that phrase reworded, Claude Opus~5.5, Sonnet~5.5 and GPT-6 Astra still read 1.29 or more with no help (two runs each). The simulated GPU was revised during the study; these three conditions ran on overlapping releases that serve the same job times.}
\label{tab:fixes}
\end{table}

\section{A Harness That Asks What Each Job Is For}
\label{sec:harness}

The harness changes two things about how an agent works. Before any job starts, the agent writes a plan that says what each job is for. After that it never starts, stops or watches a job: a program does, and wakes the agent only when a result needs judging.

\subsection{A Plan, Then Verdicts}

The plan quotes the brief's rule, names the \emph{candidates}, the options the decision is about, and lists every job the rule could need, each with a guessed length and a role. A \emph{check} is the cheapest result that could rule its candidate out. A \emph{needed} job is the rest of what the rule requires, and it waits until its candidate's checks have passed. If the rule reads a value mid-run, the plan names its log line; if it can be settled before every run is in, the plan writes the interim looks.

\begin{figure}[t]
\centering
\includegraphics[width=\columnwidth]{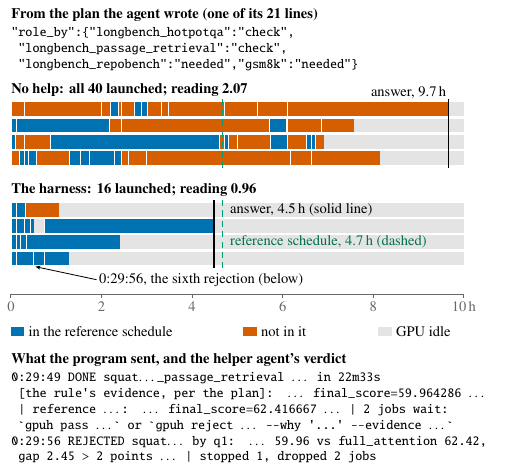}
\caption{Claude Sonnet~5.5 on the KV acceptance gate, with no help and with the harness. One row per GPU; a block is one evaluation, coloured by whether the reference schedule also runs it (job by job, not the stored shares of GPU time). Top: part of one line of the plan, which makes two short workloads each quantisation method's checks. Bottom: what the program sent a helper agent when one check finished, and the helper's verdict; an ellipsis marks a cut, times are since the agent's start, and ``reference'' there is the uncompressed model. The plan lists 36 of the 40 evaluations: it leaves out four on a workload the uncompressed model does not run. The harness run reads below 1 because the reference schedule starts the last evaluation eleven minutes later. Both runs answered correctly. The no-help run is one of the model's two with a reading (1.86 and 2.07); the harness run is one of six, whose median is 0.97.}
\label{fig:onerun}
\end{figure}

Figure~\ref{fig:onerun} follows one run of the KV acceptance gate, whose rule admits a method only within 2 points of the uncompressed model on every workload. Claude Sonnet~5.5 made two short workloads each method's checks, started the program and four helper agents, and waited. Thirty minutes in, one helper received a line: its candidate's score, 59.96, beside the uncompressed model's 62.42. Seven seconds later it rejected the candidate. The program stopped one running evaluation and dropped two queued, one the 8.1-hour evaluation of Section~3. Six candidates went this way within half an hour, taking 21 of the plan's 36 evaluations. The run answered in 4.5 hours, a reading of 0.96; with no help the same model launched all 40 and took 9.7 (2.07).

The program is 3,634 lines of Python. It gives each free GPU the runnable job that heads the longest chain of work still to come, with a check counted as heading everything it gates; a job long enough to set the finish goes first. It corrects the plan's guessed lengths from the pace jobs print in their logs. It never applies the rule: it sets a candidate's number beside the one it is judged against, and an agent decides. It needs a coding agent with background commands that wake it, helper agents with their own context, and messages between them.

\subsection{Results and Cost}

We ran the harness with Claude Haiku~5.5, Sonnet~5.5 and Opus~5.5 in Claude Code: 179 runs, 174 with a supported answer, hence a reading; the other five, Haiku's, were handed back or ended before their evidence was in. Of the 26 model-and-task pairs with a no-help reading, it improves 25, and pairs with a typical reading of 1.25 or more fall from 25 to 3. The excess sums each pair's typical reading above 1.0, a reading below 1.0 counting as zero. It falls from 23.3 to 2.5, about 90\% less (86\% without many-candidates adoption), and pools readings against different comparison schedules. Of the 2.5 left, precision stopping holds 1.06, the 8-GPU table 0.60, many-candidates adoption 0.46, data filtering 0.20, and the other five tasks 0.13; their fifteen pairs read 0.92 to 1.02.

The 25 no-help pairs at 1.25 or more are there by selection: the tasks were kept for their no-help waste (Section~3). The harness played no part in that choice, but it was tuned on all nine. On five tasks its runs used a later release of the simulated GPU than most no-help runs; each release had to serve the same job times as the one before.

In the \emph{written procedure}, an earlier version, agents carried out such a plan by hand; it is not the harness minus its program. It ran on three tasks with Haiku and Sonnet, three runs a pair. For one early rejection, words were enough: 1.02 for both on optimizer adoption~A, against 1.02 and 1.01. For forty jobs packed over nine hours they were not: 1.27 and 1.13 on the KV results table with the plan given, against 1.00.

By pair, a typical harness run costs \$0.17 to \$0.42 (Haiku), \$2.28 to \$7.16 (Sonnet) or \$4.60 to \$14.89 (Opus) in tokens. On 25 pairs it frees 1.2 to 57 GPU-hours a run, \$8 to \$394 at \$6.88 per H100-hour: allocated GPU-hours computed from replayed timings, not a GPU bill.

\subsection{Where It Stops}

\begin{figure}[t]
\centering
\includegraphics[width=\columnwidth]{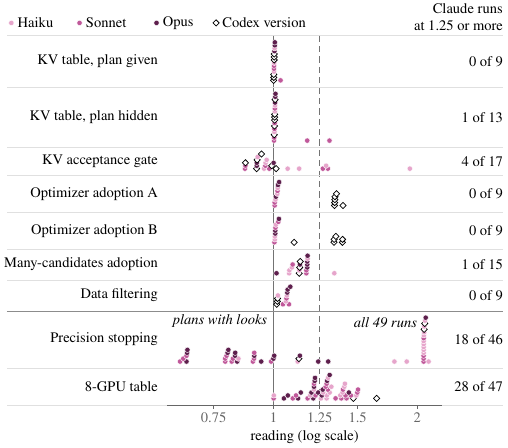}
\caption{Every harness run with a reading: 174 runs by three Claude models in Claude Code, and 41 by GPT-6 Astra and GPT-6.1 Sol with the version for Codex (open marks; a separate condition, never pooled). Marks at the same reading are piled upward. The dashed line is 1.25. The two tasks below the rule were run 48 times each and the other seven 9 to 17 times. Readings on the three results tables are against the best computed or best feasible schedule and on the other six tasks against the reference schedule, so compare within a row. Runs read below 1 on precision stopping where an agent's own test settled the question on fewer training runs than the reference schedule's procedure, and on the gate for the reason given in Figure~\ref{fig:onerun}. Many-candidates adoption sits right of 1 partly because its reference schedule knows which candidate passes (Table~1, in Section~3).}
\label{fig:everyrun}
\end{figure}

A median forgives, so Figure~\ref{fig:everyrun} shows every run. On seven tasks, 6 of 81 runs read 1.25 or more. The other two, which we ran most, hold 46 of the 52 late runs. Each shows something a plan can fail to say.

The first is when to stop. Precision stopping offers 49 training runs, and the reference schedule stops after 24. The program ends a run early only at a look the plan wrote. Of 48 plans, 27 wrote looks (Opus 15 of 16, Sonnet 12 of 16, Haiku none), and their runs read 0.64 to 1.30. Seventeen ran all 49 and read 2.06 to 2.11 where they have a reading; nine of them chose one final analysis, in one plan's words ``because the 95\% simultaneous interval needs the full sample''. The other four stopped short with no look: a smaller fixed design (1.12) and three agents' own stops (0.85, 1.79, 1.90). A fixed design is defensible statistics; no program can add looks without choosing the experiment's design.

The second is which job is long. On the 8-GPU table one 8.1-hour evaluation sets the finish, so it must start at once. Every plan under-stated it against its own other estimates, by 1.4 to 12 times, and in 45 of 47 runs the program started it after its first wave; 28 read 1.25 or more. Sonnet's typical reading there, 1.30 against 1.26 with no help, is the one pair the harness did not improve; the 0.04 lies inside the spread of its three no-help readings (1.26 to 1.65). Part of this is the program's. Recomputed with the job's true length in each plan (a computation on recorded paces, not a new run), it starts the job at once in only 21 of 47. In 21 others its own prediction, pooled over every method's lengths on that workload, pulls the slow job down. A second fault, a speed estimate that drifts, changes little: where it ran high, it offset the plan's under-estimate.

The six late runs elsewhere have named causes, mostly gaps in the program: the uncompressed model's long evaluation waiting behind the candidates' (three gate runs, 1.27 to 1.30), equal estimates run in plan order (1.31), and a log-line pattern that never matched, with no warning (1.34). In the sixth the agent added a 3.7-hour evaluation to check a tied score (1.93). No gap is fixed: the harness was frozen.

The same program, with instructions for Codex, made 41 runs with GPT-6 Astra and GPT-6.1 Sol. Results are mixed: it removes 63\% of Astra's excess and 48\% of Sol's. Over all five models, the summed excess of the 39 model-and-task pairs falls from 31.3 to 5.6, by 82\%. A Codex run costs \$1.88 to \$11.71 in tokens with Sol and \$17.61 to \$182.62 with Astra, at list prices. Both limits recur with Astra. On optimizer adoption~A and~B, where no Claude run reads above 1.03, eleven of twelve Codex runs read 1.34 to 1.40: their plans declared the rejected setup's runs needed, citing the brief's request for ``each setup's mean''.

For a builder, the program is the cheap part. The plan is not: when an experiment has seen enough, and which job will be long, are judgements about the research, made before any result is in. The evidence covers jobs known in advance, one GPU each; experiments invented during a search are untested.

\section{Limitations and Conclusion}
\label{sec:conclusion}

\paragraph{Limitations.} This paper does not show how often research invites waste: its nine tasks were kept because their waste persisted. It does not show agents on real GPUs: our hardware check ran 194 jobs with no agent, matching one task's decision only under the recorded runs' PyTorch version. It does not show how far the harness carries: we adjusted it on all nine tasks, and on five it ran on a later release of the simulated GPU than most no-help runs, at unchanged job times. The comparison schedules know job lengths the briefs omit and, on one task, which candidate passes. Two briefs ask for each setup's mean, which can be read as asking for runs we count as unneeded; with one reworded, three models still read 1.29 or more (two runs each). Runs are few: 24 of the 26 no-help medians rest on two or three.

\paragraph{Conclusion.} We asked whether AI research agents turn the GPUs they are given into research progress, and answered by fixing the result and measuring the time. ComputeWise is nine tasks built from recorded GPU runs, a simulated GPU that replays them to an agent's code, and the reading: an agent's time to a supported answer over a comparison schedule's time.

With no help, the median of 125 readings, pooled over tasks with different comparison schedules, was 1.44. Pooled over 148 runs' GPU time, 52\% went to needed work, 22\% to unneeded work, and 26\% was idle. Keeping the GPUs busy was not the same as using them well: of the 24 runs at least 90\% busy, 16 read 1.25 or more, our waste level, fixed before any screening run was read. On the short job lists of our first benchmark, a hint sentence or a scheduler brought each wasteful model's median time under 1.25 times the best schedule of its jobs; on the nine tasks, kept only where waste survived both, neither removed it.

In the harness, the model states the evidence needed and judges results; code dispatches the jobs. With three Claude models in Claude Code it improved 25 of 26 model-and-task pairs and removed about 90\% of the summed excess of their median readings over 1.0: delay beyond the comparison schedules, not GPU-hours saved. The three pairs still at 1.25 or more start from plans that left out an interim look at results or under-stated the longest job, an error the harness's code compounded. Its Codex version removed 63\% of GPT-6 Astra's excess over nine tasks and 48\% of GPT-6.1 Sol's over four.

ComputeWise is open source. Its code holds the nine tasks; the simulated GPU, whose stand-in NVIDIA driver is built from the NVML and nvidia-smi mocks of NVIDIA's k8s-test-infra; and the harness, as a plugin for Claude Code and Codex. For the four tasks built from MLS-Bench, a script downloads MLS-Bench and builds them. Its data holds per-run metrics from our experiments (scores, times, GPU occupancy and counts, no transcripts). Models and datasets made by others are named at pinned revisions, not copied. With ComputeWise, a builder can take that reading of their own agent without a GPU. The run in Figure~\ref{fig:onetask} kept its GPUs 93\% busy and took 19.0 hours where a careful schedule takes 4.0. GPU utilisation cannot tell a run whose busy time its answer needed from one kept busy by work the answer did not need; the time to a supported answer can.

\bibliographystyle{named}
\bibliography{references}

\providecommand{\polhk}[1]{\setbox0=\hbox{#1}{\ooalign{\hidewidth\lower1.5ex\hbox{`}\hidewidth\crcr\unhbox0}}}
\begin{thebibliography}{}

\bibitem[\protect\citeauthoryear{Agrawal \bgroup \em et al.\egroup
  }{2026}]{revati}
Amey Agrawal, Mayank Yadav, Sukrit Kumar, Anirudha Agrawal, Garv Ghai,
  Souradeep Bera, Elton Pinto, Sirish Gambhira, Mohammad Adain, Kasra Sohrab,
  Chus Antonanzas, and Alexey Tumanov.
\newblock Revati: Transparent {GPU}-free time-warp emulation for {LLM} serving.
\newblock {\em arXiv preprint arXiv:2601.00397}, 2026.

\bibitem[\protect\citeauthoryear{Au \bgroup \em et al.\egroup
  }{2026}]{recovering-wasted-compute-autoresearch}
Kwok~Chun Au, Abhigyan Acherjee, Amrutha Rao, Zaiqian Chen, Kazem Meidani,
  C.~Bayan Bruss, and Micah Goldblum.
\newblock Recovering wasted compute in autoresearch agents.
\newblock In {\em Conference on Language Modeling (COLM)}, 2026.

\bibitem[\protect\citeauthoryear{Brilliantov \bgroup \em et al.\egroup
  }{2026}]{how-much-harness-mle}
Kirill Brilliantov, Alejandro Hern{\'a}ndez-Cano, and Emmanuel Abb{\'e}.
\newblock How much of a harness does a strong agent need for autonomous {ML}
  engineering?
\newblock {\em arXiv preprint arXiv:2609.40303}, 2026.

\bibitem[\protect\citeauthoryear{Chan \bgroup \em et al.\egroup
  }{2025}]{mle-bench}
Jun~Shern Chan, Neil Chowdhury, Oliver Jaffe, James Aung, Dane Sherburn, Evan
  Mays, Giulio Starace, Kevin Liu, Leon Maksin, Tejal Patwardhan, Lilian Weng,
  and Aleksander M{\polhk{a}}dry.
\newblock {MLE}-bench: Evaluating machine learning agents on machine learning
  engineering.
\newblock In {\em International Conference on Learning Representations (ICLR)},
  2025.

\bibitem[\protect\citeauthoryear{Eggensperger \bgroup \em et al.\egroup
  }{2021}]{hpobench}
Katharina Eggensperger, Philipp M{\"u}ller, Neeratyoy Mallik, Matthias Feurer,
  Ren{\'e} Sass, Aaron Klein, Noor Awad, Marius Lindauer, and Frank Hutter.
\newblock {HPOBench}: A collection of reproducible multi-fidelity benchmark
  problems for {HPO}.
\newblock In {\em 35th Conference on Neural Information Processing Systems
  (NeurIPS), Track on Datasets and Benchmarks}, 2021.

\bibitem[\protect\citeauthoryear{Garikaparthi \bgroup \em et al.\egroup
  }{2026}]{researchgym}
Aniketh Garikaparthi, Manasi Patwardhan, and Arman Cohan.
\newblock {ResearchGym}: Evaluating language model agents on real-world {AI}
  research.
\newblock {\em arXiv preprint arXiv:2602.15112}, 2026.

\bibitem[\protect\citeauthoryear{Hambardzumyan \bgroup \em et al.\egroup
  }{2026}]{aira-2}
Karen Hambardzumyan, Nicolas Baldwin, Edan Toledo, Rishi Hazra, Michael
  Kuchnik, Bassel Al~Omari, Thomas~Simon Foster, Anton Protopopov,
  Jean-Christophe Gagnon-Audet, Ishita Mediratta, Kelvin Niu, Michael
  Shvartsman, Alisia Lupidi, Alexis Audran-Reiss, Parth Pathak, Tatiana
  Shavrina, Despoina Magka, Hela Momand, Derek Dunfield, Nicola Cancedda,
  Pontus Stenetorp, Carole-Jean Wu, Jakob~Nicolaus Foerster, Yoram Bachrach,
  and Martin Josifoski.
\newblock {AIRA}$_2$: Overcoming bottlenecks in {AI} research agents.
\newblock {\em arXiv preprint arXiv:2603.26499}, 2026.

\bibitem[\protect\citeauthoryear{Hogan \bgroup \em et al.\egroup
  }{2026}]{alphalab}
Brendan~R. Hogan, Xiwen Chen, James~T. Wilson, Kashif Rasul, Adel Boyarsky,
  Thomas Kamei, Anderson Schneider, and Yuriy Nevmyvaka.
\newblock {AlphaLab}: Autonomous multi-agent research across optimization
  domains with frontier {LLMs}.
\newblock {\em arXiv preprint arXiv:2604.08590}, 2026.

\bibitem[\protect\citeauthoryear{Huang \bgroup \em et al.\egroup
  }{2024}]{mlagentbench}
Qian Huang, Jian Vora, Percy Liang, and Jure Leskovec.
\newblock {MLAgentBench}: Evaluating language agents on machine learning
  experimentation.
\newblock {\em arXiv preprint arXiv:2310.03302}, 2024.

\bibitem[\protect\citeauthoryear{Jeon \bgroup \em et al.\egroup
  }{2019}]{philly-gpu-cluster-analysis}
Myeongjae Jeon, Shivaram Venkataraman, Amar Phanishayee, Junjie Qian, Wencong
  Xiao, and Fan Yang.
\newblock Analysis of large-scale multi-tenant {GPU} clusters for {DNN}
  training workloads.
\newblock In {\em Proceedings of the 2019 USENIX Annual Technical Conference
  (USENIX ATC)}, 2019.

\bibitem[\protect\citeauthoryear{Jordan \bgroup \em et al.\egroup
  }{2024}]{modded-nanogpt}
Keller Jordan, Jeremy Bernstein, Brendan Rappazzo, {@fernbear.bsky.social},
  Boza Vlado, You Jiacheng, Franz Cesista, Braden Koszarsky, and
  {@Grad62304977}.
\newblock modded-nanogpt: Speedrunning the {NanoGPT} baseline.
\newblock Software repository,
  \url{https://github.com/KellerJordan/modded-nanogpt}, 2024.

\bibitem[\protect\citeauthoryear{Lu \bgroup \em et al.\egroup
  }{2024}]{ai-scientist}
Chris Lu, Cong Lu, Robert~Tjarko Lange, Jakob Foerster, Jeff Clune, and David
  Ha.
\newblock The {AI} {S}cientist: Towards fully automated open-ended scientific
  discovery.
\newblock {\em arXiv preprint arXiv:2408.06292}, 2024.

\bibitem[\protect\citeauthoryear{Lyu \bgroup \em et al.\egroup
  }{2026}]{mls-bench}
Bohan Lyu, Yucheng Yang, Siqiao Huang, et~al.
\newblock {MLS}-{B}ench: A holistic and rigorous assessment of {AI} systems on
  building better {AI}.
\newblock {\em arXiv preprint arXiv:2605.08678}, 2026.

\bibitem[\protect\citeauthoryear{Penedo \bgroup \em et al.\egroup
  }{2024}]{fineweb}
Guilherme Penedo, Hynek Kydl{\'\i}{\v{c}}ek, Loubna Ben~allal, Anton Lozhkov,
  Margaret Mitchell, Colin Raffel, Leandro Von~Werra, and Thomas Wolf.
\newblock The {FineWeb} datasets: Decanting the web for the finest text data at
  scale.
\newblock {\em arXiv preprint arXiv:2406.17557}, 2024.

\bibitem[\protect\citeauthoryear{Qin \bgroup \em et al.\egroup
  }{2025}]{phantora}
Jianxing Qin, Jingrong Chen, Xinhao Kong, Yongji Wu, Tianjun Yuan, Liang Luo,
  Zhaodong Wang, Ying Zhang, Tingjun Chen, Alvin~R. Lebeck, and Danyang Zhuo.
\newblock Phantora: Maximizing code reuse in simulation-based machine learning
  system performance estimation.
\newblock {\em arXiv preprint arXiv:2505.01616}, 2025.

\bibitem[\protect\citeauthoryear{Salinas \bgroup \em et al.\egroup
  }{2022}]{syne-tune}
David Salinas, Matthias Seeger, Aaron Klein, Valerio Perrone, Martin Wistuba,
  and Cedric Archambeau.
\newblock Syne {T}une: A library for large-scale hyperparameter tuning and
  reproducible research.
\newblock In {\em International Conference on Automated Machine Learning
  (AutoML)}, 2022.

\bibitem[\protect\citeauthoryear{Wijk \bgroup \em et al.\egroup
  }{2024}]{re-bench}
Hjalmar Wijk, Tao Lin, Joel Becker, Sami Jawhar, Neev Parikh, Thomas Broadley,
  Lawrence Chan, Michael Chen, Josh Clymer, Jai Dhyani, Elena Ericheva,
  Katharyn Garcia, Brian Goodrich, Nikola Jurkovic, Holden Karnofsky, Megan
  Kinniment, Aron Lajko, Seraphina Nix, Lucas Sato, William Saunders, Maksym
  Taran, Ben West, and Elizabeth Barnes.
\newblock {RE}-{B}ench: Evaluating frontier {AI} {R\&D} capabilities of
  language model agents against human experts.
\newblock {\em arXiv preprint arXiv:2411.15114}, 2024.

\bibitem[\protect\citeauthoryear{Wongpanich \bgroup \em et al.\egroup
  }{2025}]{ml-fleet-efficiency-goodput}
Arissa Wongpanich, Tayo Oguntebi, Jose Baiocchi~Paredes, Yu~Emma Wang,
  Phitchaya~Mangpo Phothilimthana, Ritwika Mitra, Zongwei Zhou, Naveen Kumar,
  and Vijay~Janapa Reddi.
\newblock Machine learning fleet efficiency: Analyzing and optimizing
  large-scale {G}oogle {TPU} systems with {ML} productivity goodput.
\newblock {\em arXiv preprint arXiv:2502.06982}, 2025.

\bibitem[\protect\citeauthoryear{Xiao \bgroup \em et al.\egroup
  }{2026}]{enpire}
Wenli Xiao, Jia Xie, Tonghe Zhang, Haotian Lin, Letian Fu, Haoru Xue, Jalen Lu,
  Yi~Yang, Cunxi Dai, Zi~Wang, Jimmy Wu, Guanzhi Wang, S.~Shankar Sastry, Ken
  Goldberg, Linxi Fan, Yuke Zhu, and Guanya Shi.
\newblock {ENPIRE}: Agentic robot policy self-improvement in the real world.
\newblock {\em arXiv preprint arXiv:2606.19980}, 2026.

\bibitem[\protect\citeauthoryear{Yarlagadda \bgroup \em et al.\egroup
  }{2026}]{maya-gpu-emulation}
Srihas Yarlagadda, Amey Agrawal, Elton Pinto, Hakesh Darapaneni, Mitali
  Meratwal, Shivam Mittal, Pranavi Bajjuri, Srinivas Sridharan, and Alexey
  Tumanov.
\newblock Maya: Optimizing deep learning training workloads using {GPU} runtime
  emulation.
\newblock In {\em European Conference on Computer Systems (EuroSys)}, 2026.

\bibitem[\protect\citeauthoryear{Ying \bgroup \em et al.\egroup
  }{2019}]{nas-bench-101}
Chris Ying, Aaron Klein, Esteban Real, Eric Christiansen, Kevin Murphy, and
  Frank Hutter.
\newblock {NAS}-{B}ench-101: Towards reproducible neural architecture search.
\newblock In {\em Proceedings of the 36th International Conference on Machine
  Learning (ICML)}, 2019.

\bibitem[\protect\citeauthoryear{Zhang \bgroup \em et al.\egroup
  }{2024}]{timearena}
Yikai Zhang, Siyu Yuan, Caiyu Hu, Kyle Richardson, Yanghua Xiao, and Jiangjie
  Chen.
\newblock {TimeArena}: Shaping efficient multitasking language agents in a
  time-aware simulation.
\newblock {\em arXiv preprint arXiv:2402.05733}, 2024.

\bibitem[\protect\citeauthoryear{Zhong \bgroup \em et al.\egroup
  }{2026}]{agent4sci}
Ming Zhong, Stacy Li, Nicholas Carlini, and Matthew Jagielski.
\newblock Automating {ML} for science: Can frontier agents climb scientific
  hills in the wild?
\newblock In {\em Advances in Neural Information Processing Systems, Track on
  Evaluations and Datasets}, 2026.

\bibitem[\protect\citeauthoryear{Zhou \bgroup \em et al.\egroup
  }{2026}]{athanor}
Yufa Zhou, Yichen Wang, Jiakun Fan, Ziqi Shu, Zhichao Liu, Anru~R. Zhang,
  Neiwen Ling, and Zijun Wei.
\newblock Athanor: An agent-native system for autonomous post-training
  research.
\newblock {\em arXiv preprint arXiv:2610.21055}, 2026.

\bibitem[\protect\citeauthoryear{Zhu \bgroup \em et al.\egroup
  }{2026}]{aos-bench}
Erle Zhu, Genrui Zhang, Keyang Hu, Mingyu Su, and Hongning Wang.
\newblock {AOS}-{B}ench: Can {LLM} agents manage dynamic workloads?
\newblock {\em arXiv preprint arXiv:2610.33178}, 2026.

\end{thebibliography}
\end{document}